%% file: main.tex
\documentclass[]{bytedance}

\usepackage[toc,page,header]{appendix}

\input{resources/packages}

\input{resources/math_macro}
\usepackage{mathrsfs}
\usepackage{adjustbox}
\usepackage{multirow}
\usepackage{multirow}
\usepackage{multicol}
\usepackage{tcolorbox}
\usepackage{changepage}
\usepackage{graphicx}
\usepackage{amssymb}
\usepackage{array}
\usepackage{bm}
\usepackage{xcolor}

\usepackage{minitoc}

\usepackage{pifont}
\definecolor{mygreen}{RGB}{112, 180, 143}
\definecolor{myred}{RGB}{242, 128, 128}
\newcommand{\methodName}{DreamCharacter-1}

\title{\methodName{}: From 3D Generative Foundation Models to Product-Ready Character Generation}

\author{
\centerline{
Weizhe Liu $^{*}$\quad
Yunjie Wu $^{*}$\quad
Xiangqian Shu \quad
Guangwei Wang \quad 
    \vspace{5pt}
} 
\centerline{
    Xiangyu Xu \quad 
    Peng Li \quad 
    Yujie Li \quad
    Hengkai Guo $^{{\dagger}}$\quad
    \vspace{-5pt}
}
}

\affiliation{Intelligent Creation Team, ByteDance}

\contribution[*]{Equal Contribution}
\contribution[\dagger]{Project Lead\vspace{1em}}

\abstract{
We present \textbf{\methodName{}}, a lightweight post-adaptation framework that calibrates pretrained 3D foundation models toward high-fidelity, production-ready 3D character generation. Building upon a 3D foundation backbone, our pipeline incorporates three task-oriented components: (1) \textbf{geometry post-training}, which enhances fine-grained surface details through geometric preference optimization; (2) \textbf{texture post-training}, which synthesizes high-resolution textures and refines the appearance of occluded regions; and (3) \textbf{inference acceleration}, which enables scalable deployment. Extensive quantitative and qualitative experiments demonstrate that \methodName{} produces visually compelling and structurally robust 3D character assets, consistently surpassing state-of-the-art character generation methods.
}

\date{\today}

\checkdata[Official Page]{\url{https://dreamcharacter-x.github.io/} }
\correspondence{Hengkai Guo (\email{guohengkai@bytedance.com})}

\begin{document}
\begin{CJK*}{UTF8}{gbsn}

\maketitle

\definecolor{chinese_red}{HTML}{8B4513}
\definecolor{english_blue}{HTML}{4169E1}

\definecolor{mycolor_blue}{HTML}{E7EFFA}
\definecolor{mycolor_green}{HTML}{E6F8E0}
\definecolor{mycolor_gray}{HTML}{ECECEC}
\definecolor{pearDark}{HTML}{2980B9}
\definecolor{lightergray}{HTML}{D3D3D3}

\begin{figure}[htbp]
    \centering
    \setlength{\tabcolsep}{2pt}
    \renewcommand{\arraystretch}{0}
    \begin{tabular}{cc}
        \includegraphics[width=0.46\linewidth,height=0.25\textheight,keepaspectratio]{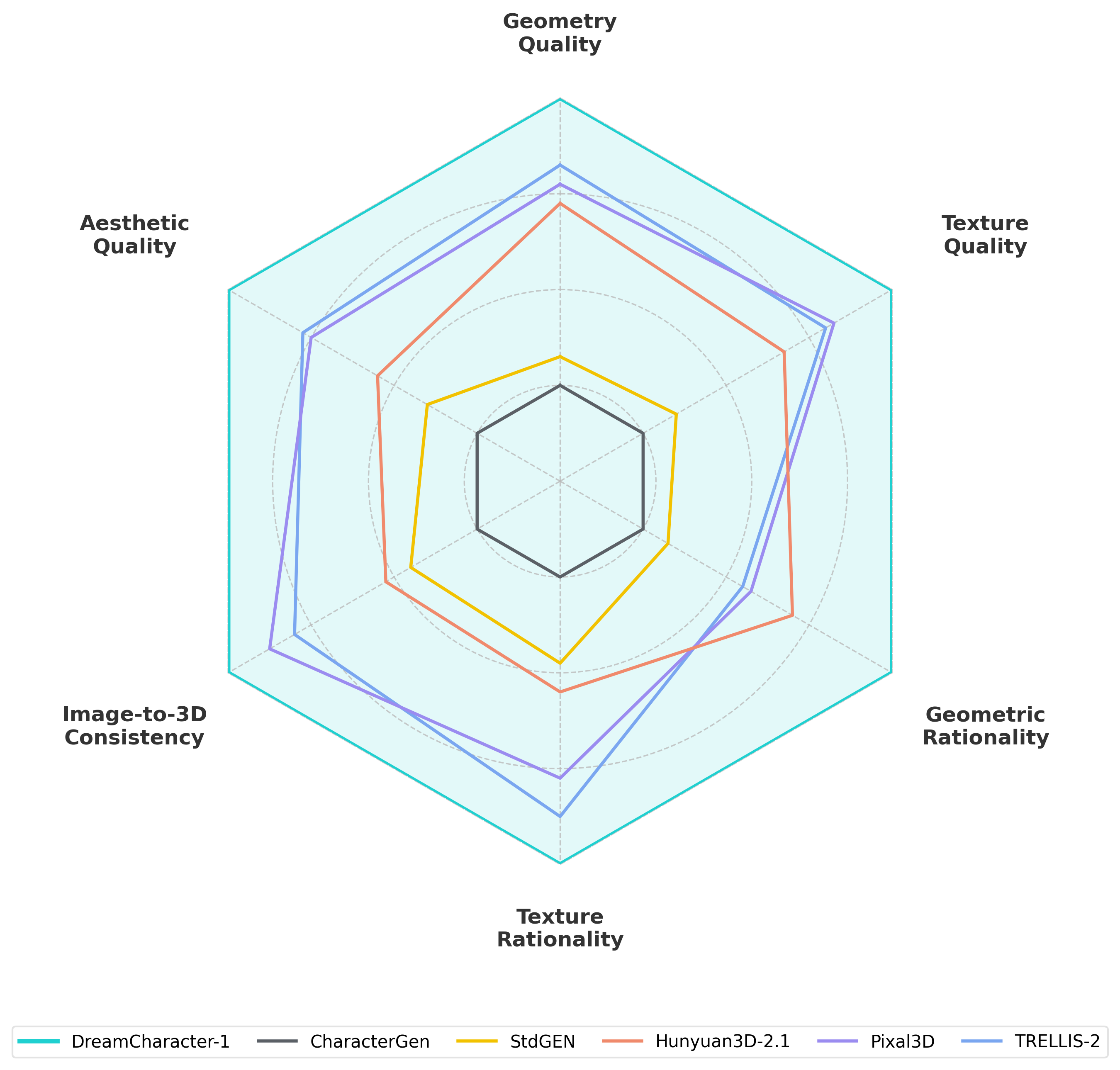} &
        \includegraphics[width=0.46\linewidth,height=0.25\textheight,keepaspectratio]{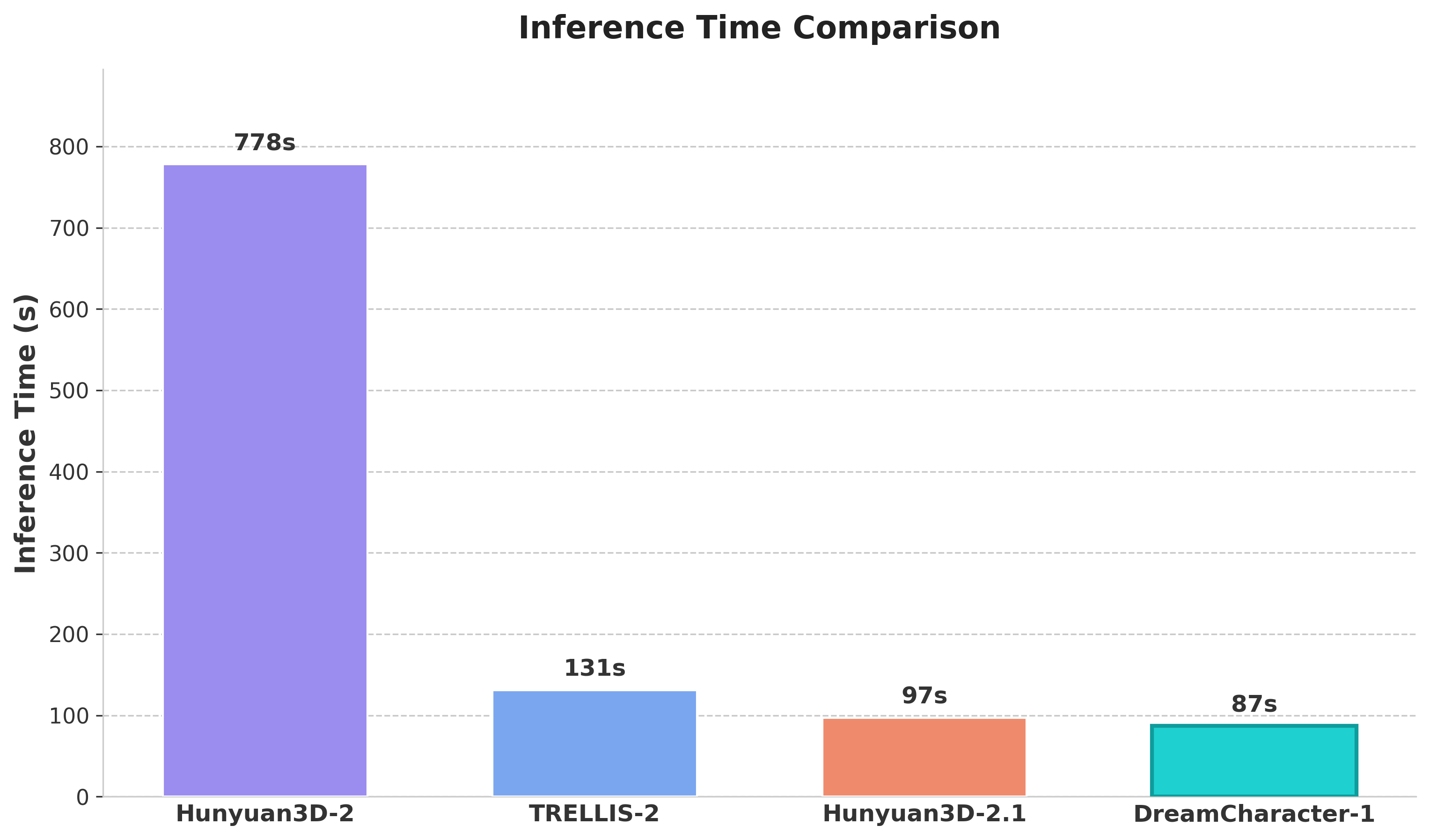} \\
    \end{tabular}
    \caption{\textbf{User study and inference efficiency.}
    Left: average human preference scores across evaluation criteria.
    Right: inference time comparison with DiT-based methods.
    \methodName{} achieves better perceptual quality and faster inference.}
    \label{fig:user_study_efficiency}
\end{figure}

\begin{figure}[htbp]
    \centering
    \includegraphics[width=0.87\linewidth]{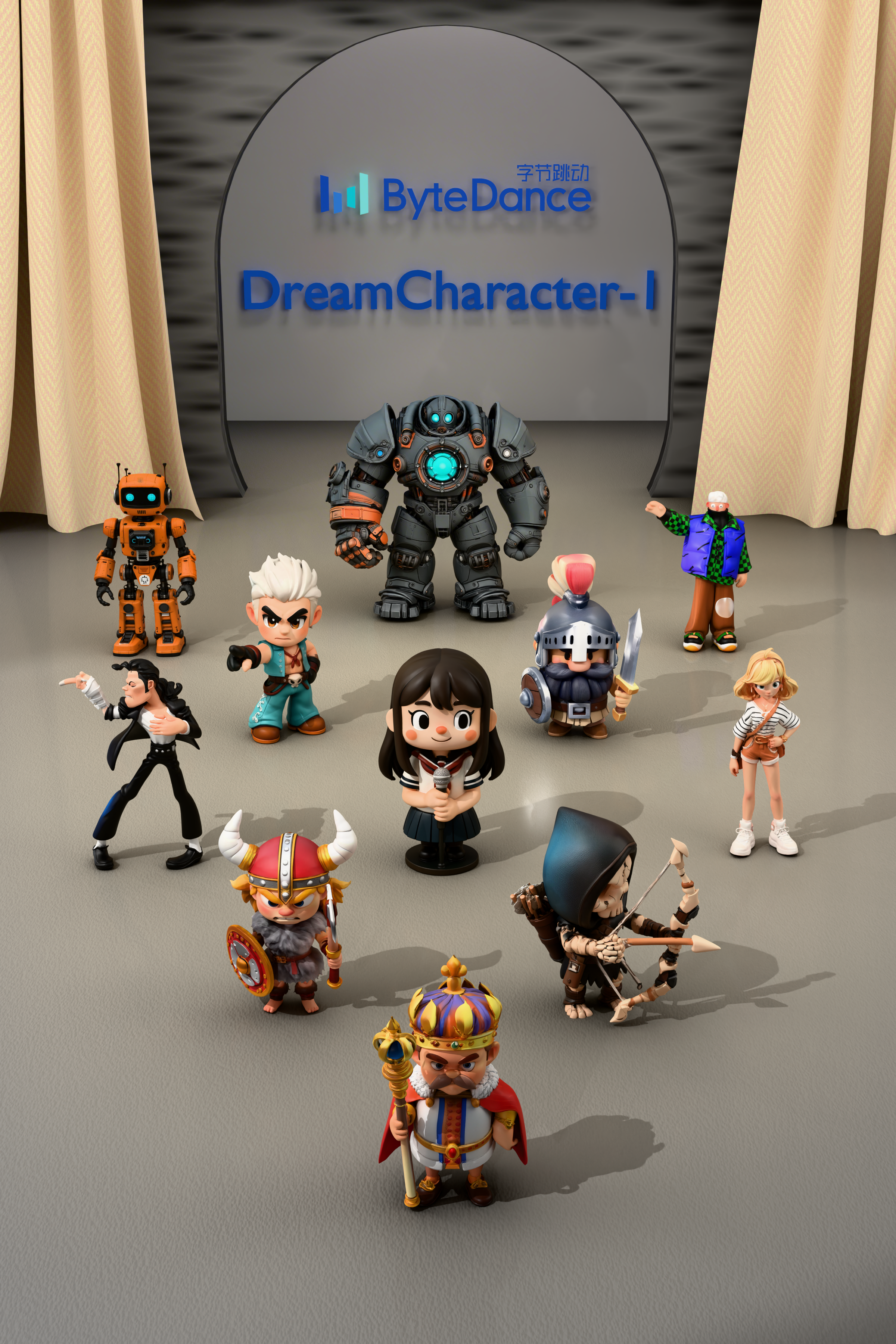}
    \captionof{figure}{\textbf{\methodName{} visualization.} Given a single reference image, our method generates high-fidelity, animation-ready 3D characters with plausible geometry, detailed appearance, and strong identity preservation.}
    \label{fig:teaser}
\end{figure}

\vspace{1em}

\clearpage

\tableofcontents

\newpage

\input{sections/introduction}

\input{sections/model}

\input{sections/training}

\input{sections/data}

\input{sections/comparison}

\input{sections/application}

\input{sections/limitation}

\input{sections/conclusion}

\input{sections/appendix}

\clearpage

\bibliographystyle{plainnat}
\bibliography{main}

\clearpage

\end{CJK*}
\end{document}

%% file: resources/packages.tex
\usepackage{natbib}

\usepackage{CJKutf8}

\usepackage{xargs}  

\usepackage{todonotes}  

\usepackage{multirow}

\usepackage{cleveref}

\usepackage{amsmath}
\usepackage{dsfont}

\usepackage{subcaption}

\usepackage{svg}

%% file: sections/introduction.tex
\section{Introduction}

General 3D asset generation~\cite{seed3d1.0, seed3d2.0, hunyuan3d2.5} has become a core problem in computer vision, computer graphics, game production, physical simulation, and embodied AI. Driven by the rapid development of pretrained generative models, modern 3D foundation models have made substantial progress in generic object reconstruction and generation, enabling flexible asset creation from monocular images, multi-view observations, and natural language prompts. Existing generic 3D pipelines can already generate daily objects, architectural models, and simple rigid assets with reasonable efficiency, significantly lowering the barrier to 3D content creation. However, current state-of-the-art 3D foundation models still fall short of producing product-ready 3D character assets, while specialized character-centric generation methods likewise remain unable to satisfy industrial production standards~\cite{peng2024charactergen, he2026stdgen++}. In practical production workflows, a deployable 3D character must satisfy a set of strict and multidimensional requirements: it must preserve identity consistency with the input reference, recover delicate high-frequency geometric details, produce high-resolution and view-consistent textures, complete plausible geometry and appearance in self-occluded or invisible regions, and remain compatible with downstream production constraints such as universal skeletal articulation and lightweight runtime performance. Compared with ordinary rigid 3D assets, elevating generated digital characters to fully product-ready quality therefore involves substantially greater unresolved technical and practical challenges.

Despite notable recent advances in character-oriented 3D generation, existing methods still fail to simultaneously achieve high visual fidelity, globally coherent structural plausibility, and production-grade robustness for industrial workflows. From the perspective of geometry, mainstream reconstruction and generation paradigms remain incapable of faithfully recovering high-frequency surface details, often producing oversmoothed meshes, incomplete thin structures, blurred sharp features, and implausible backside reconstructions~\cite{tochilkin2024triposr, li2025triposg, li2025craftsman3d}. In addition, they frequently exhibit poor aesthetic quality and anatomically implausible body proportions, which further hinders automatic rigging and downstream animation deployment~\cite{hong2024lrm, pandora3d, cui2024neusdfusion, gao2025mars}. These limitations become even more severe in human character modeling, which contains highly intricate semantic components such as irregular facial contours, layered garment folds, wearable accessories, and strand-based hair structures. As a result, generated character assets often suffer from coarse surface granularity, misaligned semantic part composition, and structurally unreasonable local geometry. Such defects make current outputs unsuitable for high-end asset pipelines, which demand both rigorous geometric accuracy and professional artistic quality.

High-fidelity texture synthesis constitutes another fundamental bottleneck for production-level character generation. In real-world acquisition scenarios, input reference images are inevitably affected by partial physical occlusion, dynamic cast shadows, uneven specular highlights, and scene-dependent ambient illumination, making it difficult for existing models to disentangle intrinsic material albedo from extrinsic lighting effects~\cite{mvadapter}. Consequently, current generative methods often produce lighting-entangled texture maps, suffer from severe cross-view inconsistency, and hallucinate unreasonable appearances in invisible regions. For identity-sensitive 3D character generation, such texture artifacts can break facial identity consistency, distort clothing semantics, and erase subtle appearance cues. These problems fundamentally undermine visual realism and prevent the generated assets from being used in demanding downstream applications, such as high-end game asset production, virtual human production, and avatar customization.

Furthermore, many methods that improve generation quality rely on enlarged backbone architectures and per-instance iterative optimization, resulting in prohibitive training cost and slow inference speed. This unfavorable trade-off between quality and efficiency further limits the integration of existing approaches into standardized and cost-effective industrial character production pipelines.

To address the intertwined challenges of geometry, texture, and industrial deployment, we present a unified post-adaptation framework, \textbf{\methodName{}}, which repurposes pretrained general-purpose 3D foundation models through task-specific post-training optimization rather than full backbone retraining. Our framework is explicitly designed to satisfy three indispensable requirements for industrial-grade 3D character production:
\begin{enumerate}
    \item \textbf{High-Fidelity Geometry.} Our method reconstructs complete high-frequency geometric details, including slender thin structures and sharp edge features, while preserving strict identity consistency with the input reference images. It also generates plausible backside geometry that is consistent with front-view observations and produces anatomically reasonable, animation-ready body topology compatible with standard rigging and downstream animation workflows.

    \item \textbf{High-Quality Texture.} Our framework synthesizes high-resolution and view-consistent texture maps that are robust to lighting artifacts and visual distractions in the input images, while recovering vivid texture details in self-occluded character regions. In addition, it produces semantic-aware UV layouts that allocate higher texture resolution to visually important regions.

    \item \textbf{Practical Efficiency.} Our method enables fast per-asset inference through optimized 3D representation design, lightweight model distillation, sparse-attention refinement, and holistic end-to-end pipeline acceleration for large-scale production scenarios.
\end{enumerate}

To fulfill these three objectives, we perform targeted optimization along two complementary branches: geometry modeling and texture modeling. For geometry generation, we adopt a hierarchical coarse-to-fine two-stage pipeline. The global coarse stage focuses on establishing a plausible overall body topology, while the local refinement stage enhances fine-grained geometric details and image--mesh identity consistency through structured voxel latent representations and multi-scale image conditioning. Moreover, we integrate multi-metric geometric reward models with reinforcement learning to jointly improve anatomical plausibility and artistic visual quality. For texture generation, we build upon a pretrained 2D image generation foundation model to produce high-resolution texture maps and strengthen cross-view texture coherence. We further introduce a dual-mesh decoupled texturing strategy together with sparse-voxel 3D inpainting to suppress unrealistic texture hallucinations in invisible and self-occluded character regions. Extensive experiments demonstrate that our approach generates visually compelling and structurally standardized 3D characters with native animation compatibility, providing a practical path toward bridging the long-standing gap between academic 3D character generation research and real-world industrial asset production.

%% file: sections/model.tex
\section{Model Design}

\subsection{Geometry}

Our geometry pipeline follows a two-stage coarse-to-fine formulation in the SDF latent space, built upon an image-conditioned latent diffusion model under the 3DShape2VecSet~\cite{3dshape2vecset} paradigm. Unlike single-stage generation schemes that must simultaneously model global structure and local geometric details, our design explicitly decouples these objectives into a coarse generation stage and a subsequent refinement stage, thereby improving both geometric fidelity and structural stability, in a manner similar to prior coarse-to-fine 3D generation frameworks~\cite{lai2025lattice, jia2025ultrashape, seed3d2.0}.

The coarse stage is initialized from a pretrained model~\cite{seed3d1.0}, which consists of a Shape-VAE~\cite{vae} and a Shape-DiT~\cite{liu2022flow}. The Shape-VAE learns a compact and expressive latent space for 3D geometry, enabling efficient encoding and decoding of continuous shape representations. On top of this latent space, the Shape-DiT, implemented as a rectified-flow-based diffusion transformer, predicts geometry latents conditioned on the input image and establishes the overall character structure, proportions, and large-scale geometric layout.

A key component of our method is the subsequent refinement stage, which takes the coarse geometry as an explicit structured condition and focuses detail synthesis on geometrically valid regions. This design allows the model to devote its capacity to recovering high-frequency surface variations without perturbing already plausible global structure. As a result, the refinement stage improves local sharpness, thin structures, and geometric consistency in challenging regions, while avoiding unnecessary modifications in unreliable or invalid areas.

\begin{figure}[t]
    \centering
    \includegraphics[width=0.9\linewidth]{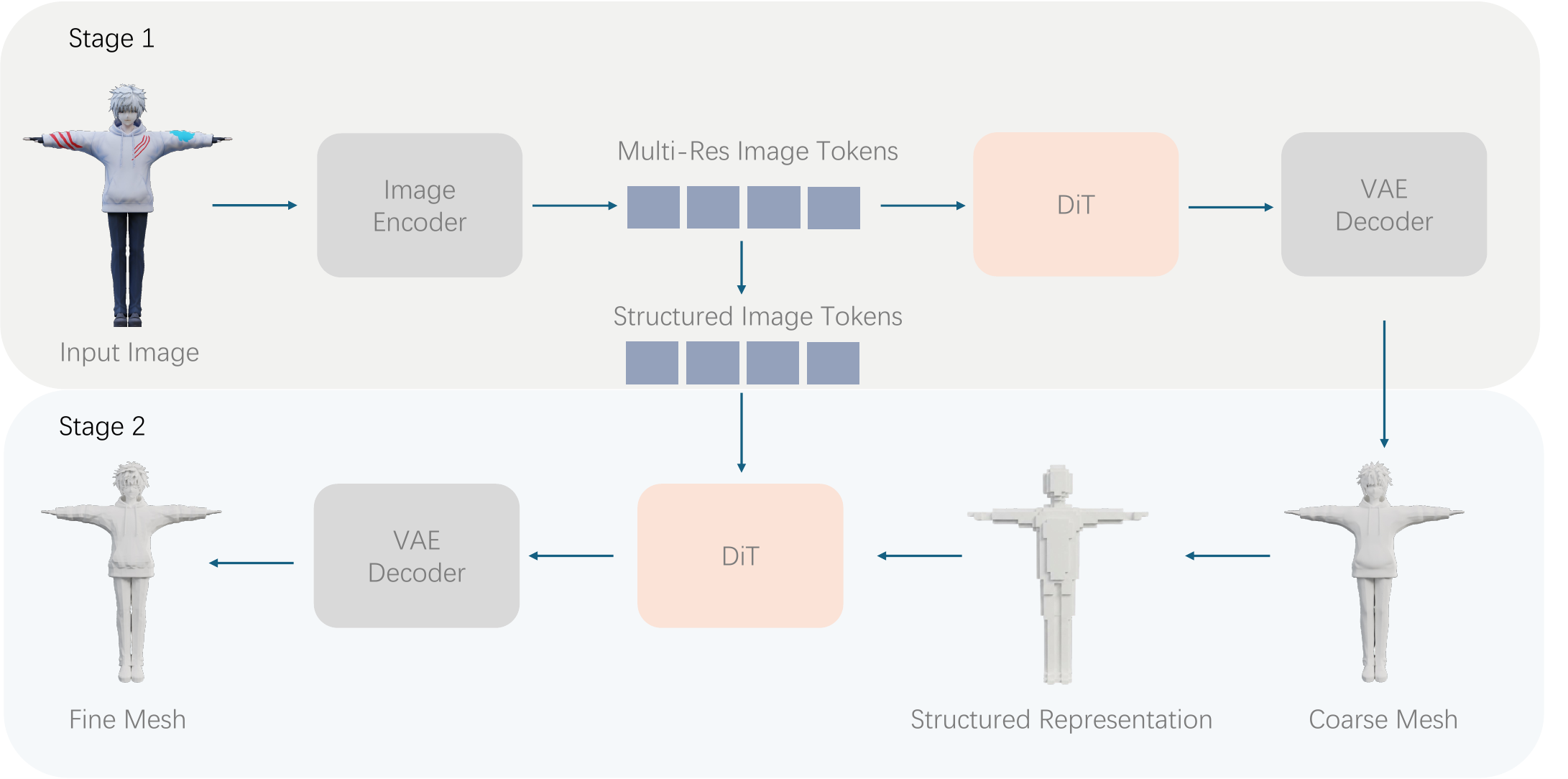}
    \caption{\textbf{Overview of the geometry generation pipeline.} In the first stage, a coarse geometric structure is generated from the input image. In the second stage, this coarse structure is further refined to produce a mesh with significantly richer geometric details.}
    \label{fig:geometry_pipeline}
\end{figure}

As shown in Figure~\ref{fig:geometry_pipeline}, the full pipeline first maps input geometry into a compact SDF latent representation, then performs image-conditioned latent prediction in a coarse-to-fine manner, and finally decodes the refined latent into the output continuous geometry field. This hierarchical generation strategy provides a stronger inductive bias for high-quality 3D character generation, yielding better detail recovery and improved structural plausibility over conventional one-stage formulations.

\subsubsection{Shape-VAE}

The Shape-VAE utilized in our coarse stage inherits the design from~\cite{seed3d1.0}, whose overall architecture resembles Dora~\cite{dora}. For the refinement stage, we first perform uniform point sampling across the full mesh surface, then conduct dense supplementary sampling over geometrically sharp regions alongside their associated surface normals. This sampling scheme yields a far denser point distribution than that adopted in the coarse stage to capture delicate local geometry. The proposed sampling strategy achieves balanced coverage of global shape layouts and local high-frequency details, which is critical for faithfully retaining sharp edges and thin geometric primitives. All sampled points are augmented with positional encodings and fused with surface normal information to learn a compact shape latent embedding.

Formally, given a collection of sampled point-normal pairs
\begin{equation}
\mathcal{P} = \{(\mathbf{x}_i, \mathbf{n}_i)\}_{i=1}^{N},
\end{equation}
where $\mathbf{x}_i \in \mathbb{R}^3$ represents a spatial sample point and $\mathbf{n}_i \in \mathbb{R}^3$ is its corresponding surface normal, the encoder transforms the input set into a condensed shape latent code via
\begin{equation}
\mathbf{z}_{\mathrm{shape}} = E(\mathcal{P}).
\end{equation}
Here, $\mathbf{z}_{\mathrm{shape}}$ acts as a compressed geometric representation that simultaneously encodes global structural layouts and subtle local surface variations.

The decoder for our refinement stage leverages self-attention to map the compact shape latent to a continuous signed distance field (SDF). Concretely, for an arbitrary spatial query point $\mathbf{q}$, we first compute its positional encoding $\gamma(\mathbf{q})$, then predict the signed distance value via
\begin{equation}
\hat{s}(\mathbf{q}) = D\bigl(\gamma(\mathbf{q}), \mathbf{z}_{\mathrm{shape}}\bigr),
\end{equation}
where $\hat{s}(\mathbf{q})$ denotes the predicted SDF value evaluated at coordinate $\mathbf{q}$. While prior work demonstrates that compact geometric representations can be formulated as unstructured vector sets~\cite{3dshape2vecset}, we empirically observe that unstructured latent embeddings deliver weaker cross-modal alignment with image guidance compared to structured alternatives~\cite{xiang2025trellis2, lai2025lattice}. Motivated by this observation, we employ a voxelized latent representation to strengthen alignment between input visual content and generated 3D shapes.

\subsubsection{Shape-DiT}

Built upon the pre-trained shape latent space, we train two separate image-conditioned latent diffusion Transformers (Shape-DiT) dedicated to the coarse generation and geometry refinement stages, respectively. The coarse-stage diffusion model follows the pipeline introduced in~\cite{seed3d1.0}, modeling the conditional distribution:
\begin{equation}
p(\mathbf{z}^{\mathrm{coarse}}_{\mathrm{shape}} \mid I),
\end{equation}
where $I$ stands for the input reference image and $\mathbf{z}^{\mathrm{coarse}}_{\mathrm{shape}}$ denotes the coarse-level shape latent code. The refinement-stage diffusion model further models a joint conditional distribution:
\begin{equation}
p(\mathbf{z}^{\mathrm{refine}}_{\mathrm{shape}} \mid I, C),
\end{equation}
where $\mathbf{z}^{\mathrm{refine}}_{\mathrm{shape}}$ is the refined shape latent and $C$ denotes the low-resolution coarse geometry output from the first stage. During training, the diffusion model learns to denoise geometry latents conditioned on both the input image and coarse geometric context, enforcing tight alignment between latent geometry features and the visual semantics of the reference image. At inference time, the refined latent code is fed into the SDF decoder to reconstruct the final high-fidelity mesh geometry.

To further boost geometric fidelity and structural plausibility within both coarse and refinement branches, we introduce a multi-scale image conditioning paradigm that exploits visual features extracted from multiple input resolutions. Specifically, low-resolution image features supply robust global semantic and structural priors, while high-resolution feature maps retain fine local appearance cues essential for recovering intricate micro geometric structures. To strengthen cross-modal alignment between input reference images and generated meshes, we further project these hierarchical image tokens into voxel space in the refinement stage. Such voxel-aligned image features are highly compatible with our structured shape latent representation, thereby enhancing fine-grained detail consistency between input visual content and the final reconstructed 3D shape.

\subsection{Texture}
\begin{figure}[t]
    \centering
    \includegraphics[width=0.9\linewidth]{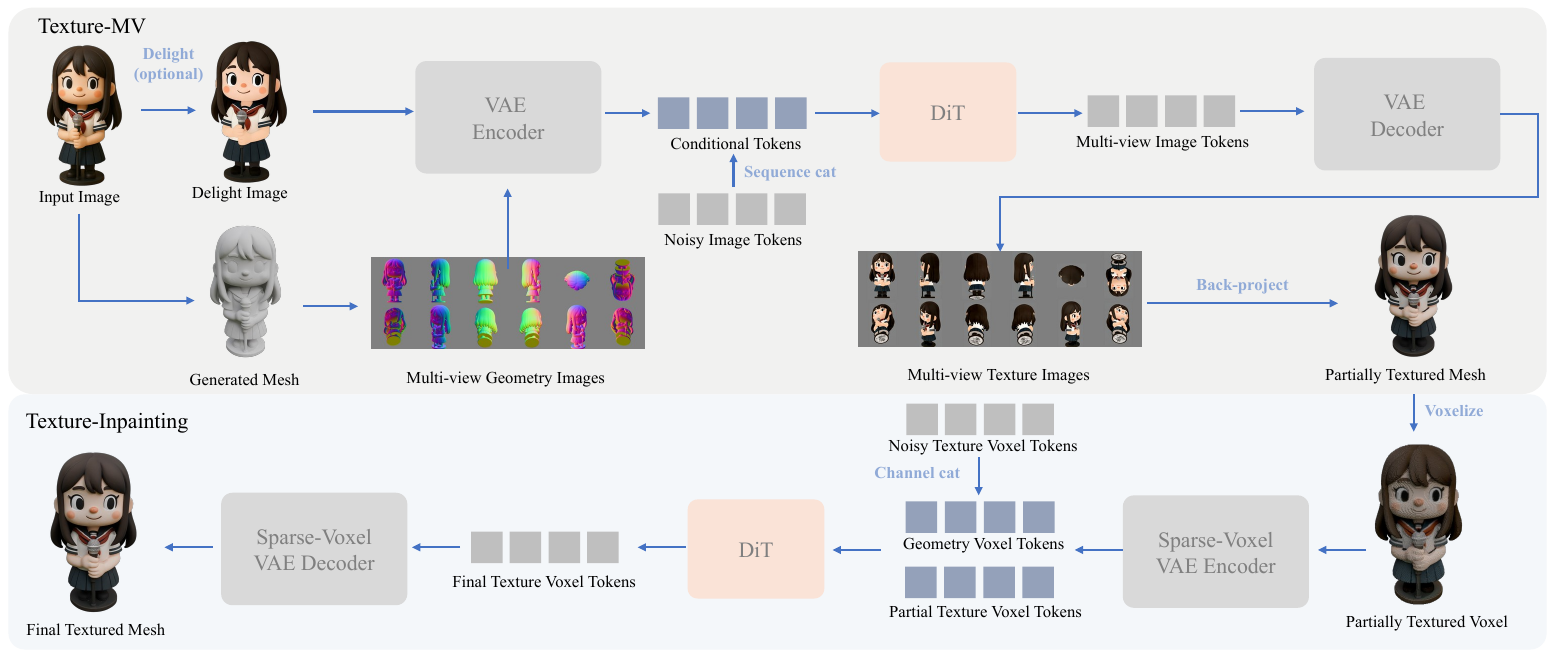}
    \caption{\textbf{Overview of the texture generation pipeline.} The Texture-MV stage generates consistent multi-view texture maps and back-projects them onto the mesh to obtain a partially textured asset. The subsequent Texture-Inpainting stage voxelizes this incomplete mesh, encodes the voxelized data as model conditioning, and predicts complete texture voxels, which are ultimately converted into the fully textured output mesh.}
    \label{fig:texture_pipeline}
\end{figure}

The complete texture generation pipeline is visualized in Figure~\ref{fig:texture_pipeline}.

Our texture framework adopts a two-stage generative paradigm. The first stage synthesizes multi-view textures within the 2D image latent space, whereas the second stage recovers missing textures for occluded regions via inpainting in a sparse voxel latent space. Concretely, the first stage is instantiated as a dual-conditional multi-view latent diffusion model composed of an image VAE and a DiT backbone. We first tile all multi-view texture snapshots into a unified high-resolution grid image, which is then compressed into a compact latent embedding via the VAE encoder. The multi-view DiT carries out flow-matching denoising over this latent space under joint conditioning of the input reference image and geometry-aware guidance maps, yielding view-consistent multi-view texture outputs. These synthesized texture planes are subsequently back-projected onto the target 3D mesh, resulting in an initially textured mesh with incomplete surface coverage caused by limited viewing angles and self-occlusion.

The second stage resolves such incomplete textures via a dedicated sparse-voxel texture inpainting module. Taking the partially textured mesh paired with its underlying geometric priors as input, the module encodes these signals into a sparse 3D voxel latent space as conditioning information, and reconstructs full texture representations through a conditional DiT-driven flow-matching denoising procedure. As summarized in Figure~\ref{fig:texture_pipeline}, the full pipeline operates sequentially: it first generates consistent multi-view texture maps with the multi-view DiT, back-projects the 2D textures onto geometry to acquire a partially covered mesh, and finally fills all occluded missing regions using the 3D inpainting DiT to output a fully textured, production-ready 3D character asset.

To further boost the industrial practicality of our system, we integrate a suite of auxiliary pre-processing and post-processing modules, including image de-lighting, semantic UV decomposition, and dual-mesh decoupled texturing. These auxiliary modules bring extra performance gains for high-quality character generation while remaining largely orthogonal to our core model training and optimization designs.

\subsubsection{Texture-MV}
Texture-MV is initialized from a pre-trained multi-view generative model~\cite{seed3d1.0} and optimized to model the conditional distribution for view-consistent multi-view texture synthesis:
\begin{equation}
p(x \mid g, i, c),
\end{equation}
where $x$ stands for the target multi-view texture images, $g$ represents spatially aligned multi-view geometric conditions including normal maps and canonical coordinate maps, $i$ refers to the input reference image, and $c$ is an optional text prompt. We first tile all multi-view conditional maps and target texture frames into a unified grid-format composite image, which is subsequently compressed into a compact latent embedding via a pre-trained image VAE. Built upon this latent representation, we adopt an in-context conditioning scheme and fit the target distribution with a Multi-Modal Diffusion Transformer (MMDiT) backbone~\cite{esser2024scaling}.

To enable sufficient cross-modal token communication, we incorporate cross-modal RoPE positional encoding~\cite{su2024roformer}. Concretely, every input token is assigned a dedicated 2D RoPE embedding. The complete input token sequence is ordered as noisy multi-view texture tokens, geometric condition tokens, reference image tokens, and text prompt tokens.

\subsubsection{Texture Inpainting}
While Texture-MV produces dense multi-view texture predictions, certain self-occluded regions, such as neck areas covered by long hair or surfaces hidden from all reference views, remain difficult to infer reliably from view-based observations alone. To address this limitation, we introduce a dedicated Texture-Inpainting module for completing textures in poorly observed or unobserved surface regions.

Recent works have extensively explored 3D native texture models~\cite{xiang2025trellis2,he2026hitem3d,zeng2026textrix,chen2026lafite}. Similarly, our Texture-Inpainting module builds upon a pre-trained 3D native texture model and adapts it to the inpainting setting. Given the partially textured mesh obtained from multi-view projection, Texture-Inpainting identifies incomplete surface regions and represents the available texture context in a native 3D manner. Instead of relying solely on image-space cues, the model conditions on both the underlying geometry and the observed texture space, allowing it to infer missing appearance while remaining spatially consistent with the target shape.

The inpainting model predicts a complete texture representation from the incomplete texture observation and geometric context, which is then decoded to recover a seamless texture over the entire surface. This design enables the pipeline to handle self-occlusions and limited-view ambiguities, improving texture completeness while preserving consistency with the visible regions.

\subsubsection{Pre-Processing and Post-Processing}

We incorporate a suite of auxiliary processing strategies to further elevate the visual fidelity and industrial practicality of the output textures.

\noindent\textbf{Image De-Lighting.} Our multi-view texture model accepts both shaded reference images and intrinsic albedo maps as input, offering flexible deployment across diverse real-world scenarios. To recover clean albedo from illuminated input photographs, we adapt an image foundation model for our de-lighting task, equipped with LoRA fine-tuning and tailored prompt optimization. Trained on synthetic renders with diverse lighting conditions and augmented with AIGC-based data, this module generalizes well across realistic and stylized image domains.

\begin{figure}[htbp]
    \centering
    \includegraphics[width=0.8\linewidth]{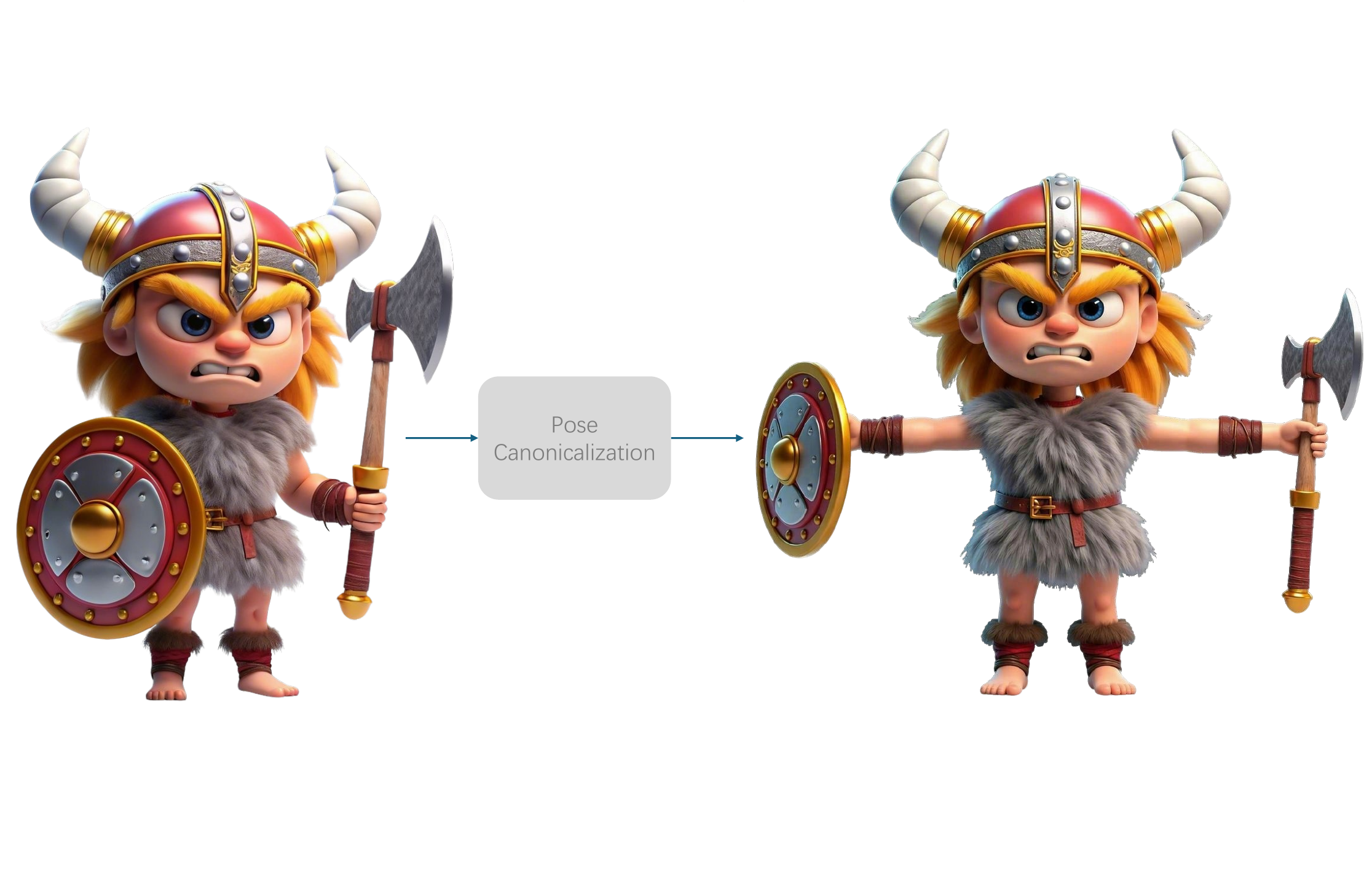}
    \caption{\textbf{Pose canonicalization results.} Characters in arbitrary poses are transformed to a canonical T-pose while preserving other attributes unchanged.}
    \label{fig:pose}
\end{figure}

\noindent\textbf{Pose Canonicalization.} To ensure stable rigging quality and reduce pose-induced occlusions, we leverage a pretrained image foundation model to transform characters from arbitrary poses into a canonical T-pose while preserving other attributes, including texture details and body proportions, as illustrated in Figure~\ref{fig:pose}.

\noindent\textbf{Decoupled Dual-Mesh Texturing.} For occluded areas containing spatially overlapping but semantically independent layers (e.g., outer clothing overlaying the human body), we split the geometry into two disjoint meshes and perform texture generation for each mesh individually. This design eliminates cross-surface texture leakage between physically separated geometry layers and enhances the visual plausibility of the final textured asset.

\noindent\textbf{Semantic UV Decomposition.} Uniform UV partitioning leads to suboptimal texel allocation, since distinct surface regions carry vastly different perceptual priorities. To address this issue, we implement semantic UV decomposition, which allocates independent UV islands with elevated texel density to visually salient semantic components. For example, facial regions, hands, decorative logos, accessories, and weapon insignias are assigned dedicated UV partitions to retain fine high-frequency details. This scheme boosts texture sharpness on perceptually critical areas while keeping the global texture resolution unchanged.

%% file: sections/training.tex
\section{Model Training and Inference}
\subsection{Geometry}
To strike a favorable balance between generalization capacity and generation fidelity, we adopt a progressive multi-stage training scheme that leverages training datasets with gradually increasing complexity and visual quality. We first train on large-scale generic 3D corpora to establish robust universal shape priors, then conduct supervised fine-tuning on artist-curated high-quality character assets to boost fine-grained visual fidelity. To enhance the plausibility of unobserved surfaces, especially rear-side geometry, we introduce explicit back-view conditioning for both training and inference pipelines. Furthermore, to elevate aesthetic quality and animation compatibility, we propose a latent-space reinforcement learning geometry optimization pipeline that incurs zero extra inference overhead. Lastly, we develop a suite of complementary acceleration techniques to support large-scale industrial deployment.

\subsubsection{Training-Time Geometry Improvements}
We incorporate multiple training-time augmentations into the geometry generation branch to lift geometric fidelity, topological rationality, and robustness against lighting and texture variations.

\noindent\textbf{Continued Training at Higher Resolution.}
Models trained under low SDF resolutions can reliably capture coarse global structures yet struggle to reconstruct delicate geometric features including sharp creases, tiny decorative ornaments, and subtle surface undulations. To tackle this limitation, we continue training the pre-trained base model with elevated SDF resolution paired with high-resolution image conditioning. This setup empowers the SDF decoder to represent rich high-frequency geometric details while inheriting stable optimization dynamics from the low-resolution pre-training phase.

\noindent\textbf{High-Quality Supervised Fine-Tuning.}
Following large-scale continued training, we perform supervised fine-tuning on a filtered high-quality asset subset featuring elaborate geometric details, clean mesh topology, and consistent alignment between geometry and rendered appearance. This stage further refines the sharpness, structural coherence, and overall visual quality of generated character meshes.

\noindent\textbf{Back-View Conditioning.}
Single-view 3D reconstruction inherently suffers from ambiguity on invisible surfaces. To generate plausible rear-side geometry, we introduce a dedicated back-view conditioning branch. This branch accepts a real back-view reference image if available; otherwise, it relies on a learned image completion module to synthesize a plausible back-view visual prior. By feeding both front-view observations and rear-side visual cues into the geometry diffusion model, our method mitigates typical reconstruction artifacts such as collapsed structures, over-flattened surfaces, and anatomically irrational back geometry. This design delivers prominent gains for characters with intricate rear-side geometry, including long hair, capes, backpacks, and asymmetric accessories.

\noindent\textbf{Reinforcement Learning for Aesthetics and Drivability.}
We further refine geometry generation via multi-objective reinforcement learning (RL), jointly optimizing perceptual visual quality and downstream animation usability. We build task-specific reward models centered on human-character visual attributes: anatomical body proportions, facial contour identity consistency, global silhouette integrity, native rigging readiness, and cross-modal image-mesh alignment. Reward signals are sourced from two complementary streams: human subjective preference annotations, and hybrid model evaluators built atop differentiable rendering pipelines and fine-tuned vision-language models (VLMs). Equipped with these multi-aspect reward functions, we conduct unified joint RL optimization. All attribute-aware reward models are integrated into a single training loop to simultaneously optimize diverse geometric properties. In contrast to separate single-attribute optimization, our joint RL workflow cuts total computational costs and alleviates inherent performance trade-offs where optimizing one attribute degrades others.

\subsubsection{Inference Acceleration}
To enable practical industrial deployment, we boost inference efficiency via two orthogonal acceleration strategies targeting different pipeline stages.

\begin{enumerate}
    \item \textbf{Student-Model Distillation}~\cite{yin2311one}.
    We distill the full multi-step diffusion teacher into a lightweight student model to drastically reduce the number of denoising steps required at inference. The student is trained to mimic the teacher's multi-step denoising trajectory, yielding accurate latent predictions in far fewer iterations. This alleviates both the iterative sampling cost and the per-sample latent prediction overhead, while preserving reconstruction quality.

    \item \textbf{Fast Mesh Extraction}.
    For high-resolution mesh reconstruction, we introduce a fast mesh extraction scheme into the voxel-space generation stage. By optimizing the reconstruction logic, it accelerates volumetric SDF decoding and enables efficient recovery of dense, high-fidelity geometry from voxel latent representations. This optimization is essential to our pipeline, as direct high-resolution 3D reconstruction would otherwise incur prohibitive memory consumption and runtime.
\end{enumerate}

\subsection{Texture}
To produce photorealistic, consistent texture maps, we design dedicated post-training refinement pipelines for both the multi-view texture foundation model and sparse voxel 3D texture inpainting model. For the multi-view branch, we conduct targeted post-training on carefully filtered high-quality data. Compared with the original base model, our refined Texture-MV delivers faster DiT attention computation, denser multi-view outputs, and stronger reference-image control. For the texture inpainting branch, we adapt input modalities under minimal parameter modification, transforming the original general-purpose 3D generation model into a texture completion module while maximally preserving pre-trained geometric and appearance priors. We also implement multiple inference acceleration schemes to guarantee stable, low-latency industrial deployment.

\subsubsection{Post-Training for Texture-MV}
We deploy a series of post-training optimizations to elevate the output quality and reference controllability of Texture-MV.

\noindent\textbf{Dense Views.}
We post-train Texture-MV with denser multi-view supervision than the baseline model~\cite{seed3d1.0}. Our training dataset contains multiple distinct camera views, substantially denser than the original multi-view setup. A denser view set suppresses artifacts on occluded surfaces and facilitates richer cross-view information exchange within the MMDiT backbone, strengthening global multi-view texture consistency. During each forward pass, we retain the positional embeddings of the base views and append dedicated positional embeddings for the additional dense views.

\noindent\textbf{Back-View Conditioning.}
The vanilla multi-view base model enforces strong supervision on the frontal viewpoint, while all other views are inferred purely from learned model priors. In contrast, Texture-MV supports up to two input reference images captured from distinct viewpoints, distinguished by independent RoPE coordinate offsets. This functionality enables precise control over character back and side appearances. Our training corpus incorporates paired front-view and back-view references; during training, each reference branch is randomly masked out with a fixed probability, encouraging the model to infer full multi-view textures conditioned on either single input.

\subsubsection{Post-Training for Texture-Inpainting}
We build the Texture-Inpainting module by adapting a pre-trained general 3D texture model through a lightweight post-training procedure. The goal is to transfer the generative prior of the original 3D texture model to the 3D inpainting setting.

To enable texture completion under partially observed inputs, we introduce the observed partial texture as an additional spatially aligned condition. This condition is represented in a 3D voxel manner consistent with the original 3D texture formulation, allowing the model to leverage existing correspondences between geometry and surface appearance. During post-training, we construct incomplete-texture cases from complete textured assets, encouraging the model to infer missing appearance regions from the surrounding valid texture and the underlying geometry. The condition is injected along the channel dimension, so that the pre-trained texture prior can be largely preserved.

\subsubsection{Inference Acceleration}
We further optimize the end-to-end texture pipeline with four complementary acceleration techniques:

\begin{enumerate}
    \item \textbf{Efficient Attention Computation.}
    To handle the heavy computation in multi-view generation, Texture-MV adopts a strategy similar to KV-cache that skips redundant attention computations, particularly those involving background or conditional tokens. This significantly reduces the computation of each DiT layer and the overall inference latency, without degrading generation quality.

    \item \textbf{Model Distillation.}
    We take our fully post-trained Texture-MV as the teacher model and train a low-NFE student model via distillation loss, removing classifier-free guidance to cut sampling overhead.

    \item \textbf{Pipeline Parallelism.}
    We split the full texture generation workflow into discrete stages: semantic UV unwrapping, geometric guidance rendering, multi-view texture generation, mesh back-projection, and sparse voxel texture inpainting. Although the overall pipeline contains inherent sequential dependencies, we parallelize stages that do not have logical ordering constraints, such as semantic UV unwrapping and geometric guidance rendering, by dispatching them to separate hardware devices or worker threads. This enables better hardware utilization and reduces end-to-end latency.
\end{enumerate}

All above optimizations drastically lower inference latency, while users can still enable full high-quality refinement when generation fidelity is prioritized over speed.

%% file: sections/data.tex
\section{Data}
The performance of image-conditioned 3D generative models is heavily governed by both the scale and quality of training data. Raw 3D assets collected from disparate heterogeneous sources suffer from severe inconsistencies and pervasive defects, including fragmented file formats, corrupted meshes, fragile thin-shell geometry, irregular topology, and low-fidelity texture maps. To resolve these limitations, we build an end-to-end automated 3D preprocessing pipeline paired with scalable data infrastructure, which converts diverse raw asset collections into standardized, high-quality training datasets stable for high-fidelity 3D character generation.

\begin{figure}[t]
    \centering
    \includegraphics[width=1.0\linewidth]{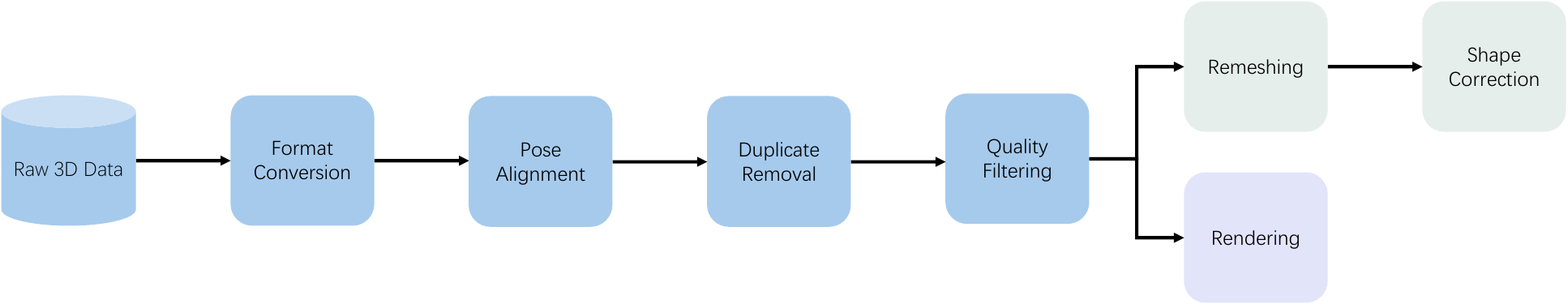}
    \caption{\textbf{Overview of our data preprocessing pipeline.} Raw 3D assets are first unified into a single standard format and aligned to canonical human pose. Duplicate and low-quality samples are filtered out, followed by dedicated geometry and texture refinement stages.}
    \label{fig:data_preprocess_pipeline}
\end{figure}

\subsection{Unified Preprocessing Pipeline}
To produce clean, standardized 3D training data and guarantee stable downstream model training, we design a systematic preprocessing workflow whose full procedure is visualized in Figure~\ref{fig:data_preprocess_pipeline}.

\noindent\textbf{Format Conversion.}
Raw 3D assets originate from multiple platforms and adopt a wide spectrum of proprietary file formats, each requiring customized parsing logic. Such format heterogeneity introduces heavy engineering overhead and frequently triggers runtime crashes during large-scale data processing. For uniform pipeline execution, we batch-convert all input assets into GLB, a lightweight format balancing compact storage, broad cross-tool compatibility, and convenient downstream geometric/texture manipulation.

\noindent\textbf{Canonical Pose Alignment.}
Canonical orientation alignment is a critical prerequisite for robust character generation, as most training assets feature front-facing human figures. For each asset, we render multi-view snapshots under fixed camera trajectories and feed these images into a pre-trained orientation classifier to predict its standard upright pose. The asset is then spatially transformed to this unified canonical coordinate system. This normalization enforces consistent spatial correspondence across geometrically similar characters, regularizes the overall data distribution, and stabilizes model convergence during training.

\noindent\textbf{Multi-Modal Deduplication.}
Aggregated 3D datasets inevitably contain duplicate or near-duplicate assets, which over-represent certain categories and shrink the effective visual and geometric diversity of the training corpus. We perform deduplication by jointly leveraging deep visual embeddings extracted from rendered views and explicit geometric shape descriptors computed from mesh surfaces. Combining appearance and geometric features makes the deduplication robust against replicas sharing identical textures, identical mesh topology, or both. All samples exceeding a predefined similarity threshold are marked as redundant and discarded.

\noindent\textbf{Hierarchical Quality Filtering.}
We conduct multi-stage quality filtering combining handcrafted geometric statistics and learned VLM-based quality assessment. First, rule-based filters eliminate low-quality assets via measurable mesh attributes such as face count and native texture resolution. Next, a fine-tuned vision-language model annotates each retained asset with comprehensive attribute tags covering aesthetic score, geometric integrity, texture fidelity, semantic category, and artistic style. These structured labels support flexible data partitioning across different training phases; high-quality subsets are isolated for high-resolution continued training and supervised fine-tuning to boost output realism.

\subsection{Geometry Preprocessing}
The geometry preprocessing branch must simultaneously preserve fine geometric details and guarantee animation-ready mesh quality. A core pain point of naive preprocessing pipelines is the accidental removal of delicate surface structures essential for industrial character production. Additionally, watertight closed meshes are mandatory for reliable SDF sampling, while thin-shell geometry easily generates unwanted double-layer artifacts during UDF-based remeshing.

To address these challenges, our geometry processing pipeline is built around three core design goals: retaining intricate geometric details throughout remeshing, detecting and repairing double-layer surfaces induced by thin geometric shells, and strengthening generalization across concept art, stylized illustrations, and other non-photorealistic reference inputs.

\subsubsection{GPU-Accelerated Remeshing}
We implement a CUDA-accelerated remeshing pipeline that strips redundant internal geometry while fully preserving exterior fine-grained surface features. The pipeline relies on unsigned distance field (UDF) representation, with final high-resolution meshes extracted via Dual Marching Cubes~\cite{nielson2004dual}. This GPU-parallelized design enables large-batch scalable processing without sacrificing surface fidelity required for precise geometric supervision signals.

\subsubsection{Thin-Shell Shape Correction}
Thin-shell geometry frequently produces overlapping double-layer surfaces after remeshing. To mitigate this artifact, we densely sample SDF values surrounding mesh surfaces and identify double-layer regions by analyzing local signed distance distribution patterns. Once problematic areas are flagged, we adaptively expand the effective surface thickness according to local SDF statistics and re-run the remeshing pass. This automated correction thoroughly suppresses double-layer mesh defects.

This repair mechanism improves latent geometry consistency and reduces supervision ambiguity during training. It is especially vital for animatable character assets, as thin overlapping body geometry would otherwise trigger rigging breakdowns and unstable skeletal deformations in downstream animation workflows.

\subsubsection{Stylized Rendering Augmentation}
We render each processed mesh from evenly distributed multi-view camera angles and augment the conditioning images via generative stylization based on Seedream~\cite{seedream2025seedream}. This augmentation exposes the geometry model to an expanded visual manifold of shading styles and texture appearances while keeping the underlying ground-truth mesh unchanged. Consequently, the model gains stronger robustness against stylized concept sketches, hand-drawn illustrations, and other non-photorealistic input references.

\subsection{Texture Preprocessing}
\subsubsection{Reference Image Normalization and Domain Adaptation}
Our model takes reference images as core conditioning signals, so we standardize input visuals to narrow the domain gap between real user inputs and training renderings. Preprocessing steps include foreground masking extraction, pose-centered cropping, and canonical camera normalization to produce geometrically consistent input imagery.

To further mitigate cross-domain appearance mismatch and enhance texture fidelity, we adopt bidirectional domain adaptation combining synthetic geometric re-rendering and generative style transfer. We translate geometry-aligned renderings toward diverse target artistic domains, while also normalizing arbitrary input images to match the internal rendering distribution seen during model training. This bidirectional adaptation unifies color tone, surface pattern, and material representation across realistic and stylized character assets.

\subsubsection{Training-Time Augmentation}
We apply diverse augmentations to both geometry and image conditioning data during model training.

\noindent\textbf{Imperfect Image-Geometry Alignment Augmentation.}
We intentionally construct training pairs with semantic consistency yet minor geometric misalignment between reference images and ground-truth meshes by perturbing reference view projections. Training on such imperfect pairs compels Texture-MV to adapt the spatial layout of generated multi-view textures to the underlying geometry while maintaining semantic alignment with the input reference image.

\noindent\textbf{Color and Geometric Jittering.}
We apply random color jitter to both reference images and ground-truth multi-view texture renders, together with independent uniform scale jitter along the X/Y/Z axes of ground-truth meshes. These augmentations expand the effective training data distribution and strengthen the generalization ability of our texture generation modules.

%% file: sections/comparison.tex
\section{Model Performance}
\subsection{Geometry Generation}

\noindent\textbf{Experimental Setup.}
Following established evaluation protocols for single-image 3D character generation~\cite{he2025stdgen,peng2024charactergen}, we adopt the benchmark dataset~\cite{chen2023panic3d} as our unified test set. We compare our approach against a suite of state-of-the-art open-source baselines: CharacterGen~\cite{peng2024charactergen}, StdGEN~\cite{he2025stdgen}, Hunyuan3D-2.0~\cite{zhao2025hunyuan3d2.0}, Hunyuan3D-2.1~\cite{hunyuan3d2025hunyuan3d2.1}, TRELLIS-1.0~\cite{xiang2025structured}, TRELLIS-2.0~\cite{xiang2025trellis2}, and Pixal3D~\cite{li2026pixal3d}. To quantify cross-modal alignment between input reference images and reconstructed meshes, we employ ULIP~\cite{xue2024ulip} and Uni3D~\cite{zhou2024uni3d} as standard geometry-image similarity metrics.

\begin{table}[t]
    \centering
    \setlength{\tabcolsep}{6pt}
    \renewcommand{\arraystretch}{1.1}
    \small
    \begin{tabular}{lcc}
        \toprule
        Method & ULIP $\uparrow$ & Uni3D $\uparrow$ \\
        \midrule
        CharacterGen      & 0.5516 & 0.6486 \\
        StdGEN            & 0.6697 & 0.7278 \\
        Hunyuan3D-2.0     & 0.8025 & 0.7968 \\
        Hunyuan3D-2.1     & 0.8273 & 0.8195 \\
        TRELLIS-1.0       & 0.8016 & 0.7953 \\
        TRELLIS-2.0       & 0.8329 & 0.8261 \\
        Pixal3D           & 0.8276 & 0.8247 \\
        \methodName{}     & \textbf{0.8532} & \textbf{0.8497} \\
        \bottomrule
    \end{tabular}
    \caption{Quantitative comparison of geometry generation metrics. Larger values indicate stronger image-mesh alignment. \methodName{} achieves state-of-the-art performance on both metrics with clear margins over all baselines.}
    \label{tab:geometry_generation_quantitative}
\end{table}

\noindent\textbf{Quantitative Results.}
As summarized in Table~\ref{tab:geometry_generation_quantitative}, \methodName{} surpasses all compared baselines on both ULIP and Uni3D. Our method yields consistent improvement against prior arts, demonstrating superior capacity to generate geometry consistent with input visual cues. The consistent improvements across two distinct cross-modal metrics verify that our reconstructed meshes exhibit tighter semantic alignment and higher structural fidelity relative to reference images.

\begin{figure}[!htbp]
    \centering
    \includegraphics[width=0.82\linewidth]{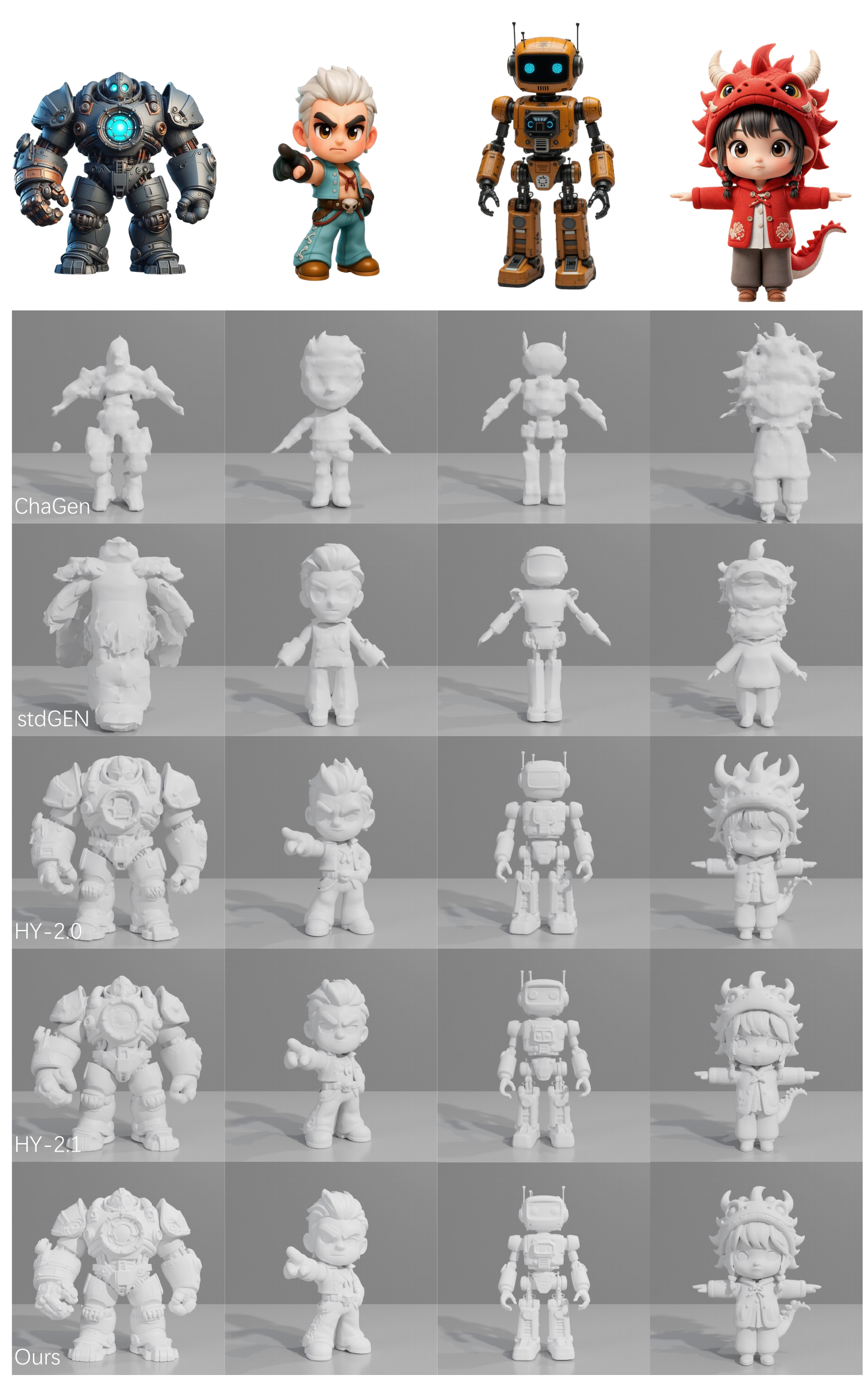}
    \caption{\textbf{Qualitative geometry generation comparison.} Raw mesh outputs from all evaluated methods are displayed. Relative to prior approaches, \methodName{} reconstructs geometry with richer micro-details and more consistent overall shapes matching the input reference image.}
    \label{fig:geometry_generation_qualitative}
\end{figure}

\noindent\textbf{Qualitative Comparisons.}
We provide side-by-side qualitative visualizations in Figure~\ref{fig:geometry_generation_qualitative} to validate that \methodName{} produces highly detailed meshes faithful to input reference images. Raw mesh renderings reveal that our method consistently outperforms all baselines in fine-detail preservation, global structural accuracy, and overall shape coherence. Unlike competing methods, our pipeline reliably recovers intricate delicate geometric features that remain challenging for existing single-view 3D generation frameworks.

\subsection{Texture Generation}

\noindent\textbf{Experimental Setup.}
We conduct texture evaluation on the same test split~\cite{chen2023panic3d}, benchmarking against widely adopted open-source 3D character generation baselines including Hunyuan3D-2.0~\cite{zhao2025hunyuan3d2.0}, Hunyuan3D-2.1~\cite{hunyuan3d2025hunyuan3d2.1}, CharacterGen~\cite{peng2024charactergen}, StdGEN~\cite{he2025stdgen}, TRELLIS-1.0~\cite{xiang2025structured}, TRELLIS-2.0~\cite{xiang2025trellis2}, and Pixal3D~\cite{li2026pixal3d}. To measure texture reconstruction quality, we render each textured mesh under a canonical front-view camera matching the input reference viewpoint and compute standard image quality metrics between rendered outputs and reference images: SSIM~\cite{wang2004image}, LPIPS~\cite{zhang2018unreasonable}, FID~\cite{heusel2017gans}, and CLIP similarity~\cite{radford2021learning}.

\begin{table}[t]
    \centering
    \setlength{\tabcolsep}{6pt}
    \renewcommand{\arraystretch}{1.1}
    \small
    \begin{tabular}{l cccc}
        \toprule
        Method & SSIM $\uparrow$ & LPIPS $\downarrow$ & FID $\downarrow$ & CLIP-Sim $\uparrow$ \\
        \midrule
        CharacterGen           & 0.8874 & 0.1876 & 0.0798 & 0.9183 \\
        StdGEN                 & 0.9000 & 0.1809 & 0.1606 & 0.8874 \\
        Hunyuan3D-2.0          & 0.9025 & 0.1159 & 0.1779 & 0.9152 \\
        Hunyuan3D-2.1          & 0.9151 & 0.0910 & 0.0678 & 0.9440 \\
        TRELLIS-1.0            & 0.9030 & 0.1193 & 0.1838 & 0.9165 \\
        TRELLIS-2.0            & 0.9188 & 0.1022 & 0.0864 & 0.9073 \\
        Pixal3D                & 0.9074 & 0.1173 & 0.0339 & 0.9231 \\
        \methodName{}          & \textbf{0.9349} & \textbf{0.0686} & \textbf{0.0319} & \textbf{0.9576} \\
        \bottomrule
    \end{tabular}
    \caption{Quantitative texture generation benchmark. All metrics compare front-view renderings of predicted textured meshes against reference images. \methodName{} establishes new state-of-the-art results on all metrics.}
    \label{tab:texture_generation_quantitative}
\end{table}

\begin{figure}[!htbp]
    \centering
    \includegraphics[width=0.82\linewidth]{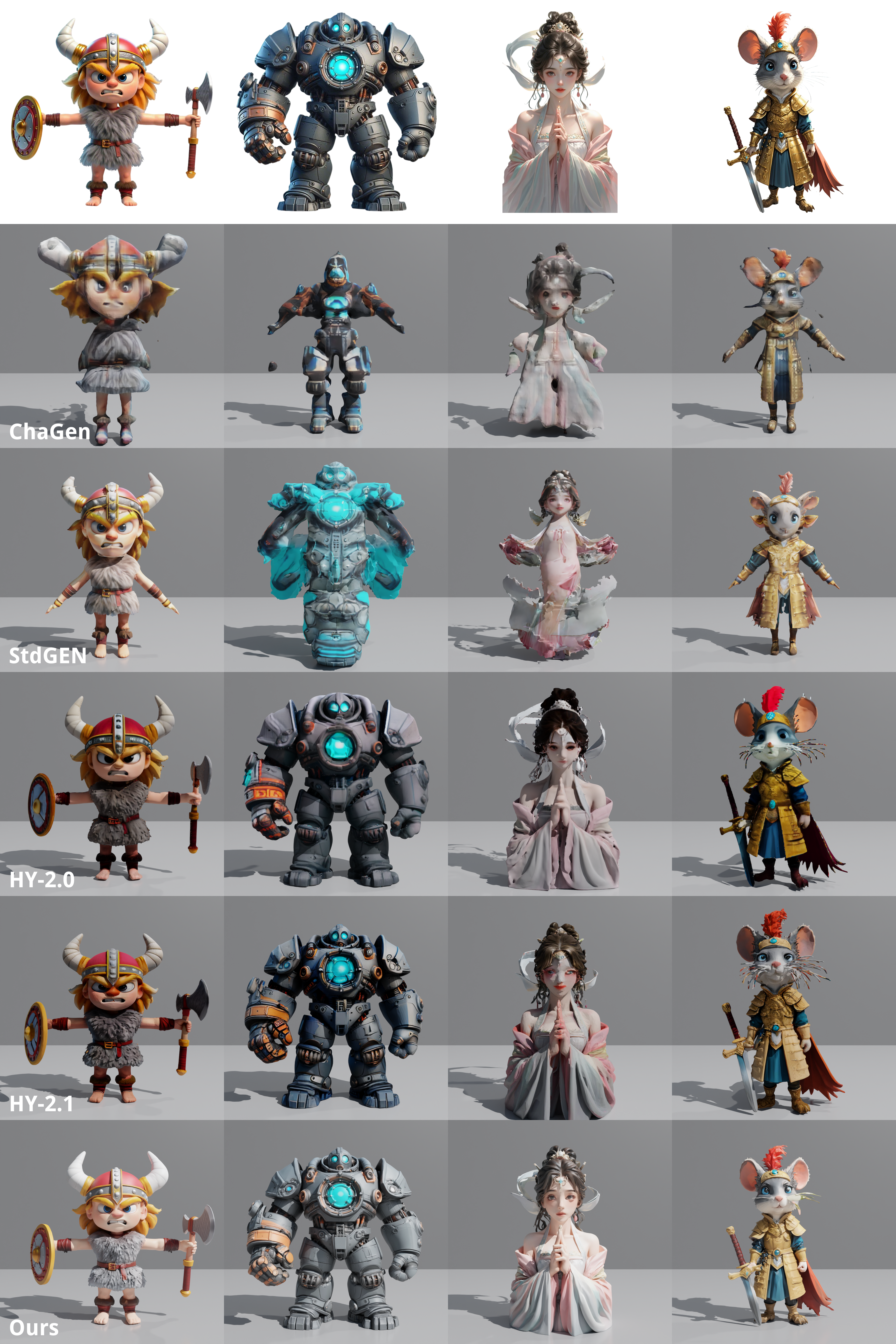}
    \caption{\textbf{Qualitative texture comparison.} Textured mesh outputs from all baseline methods are displayed side-by-side. \methodName{} generates textures with richer micro-details, accurate color reproduction, and consistent appearance across viewing angles, matching the visual characteristics of input reference images more faithfully.}
    \label{fig:texture_generation_qualitative}
\end{figure}

\noindent\textbf{Quantitative Results.}
Table~\ref{tab:texture_generation_quantitative} demonstrates that \methodName{} outperforms all baselines across every metric by a noticeable margin, achieving superior texture fidelity and cross-view consistency, verifying the flexibility and robustness of our two-stage texture generation design.

\noindent\textbf{Qualitative Comparisons.}
Figure~\ref{fig:texture_generation_qualitative} presents visual comparisons to illustrate the advantages of our texture pipeline. Visually inspecting textured mesh outputs, we observe that \methodName{} consistently surpasses baseline methods in color accuracy, fine texture detail retention, and holistic appearance consistency. Our framework reliably reproduces subtle surface patterns and maintains uniform texture appearance across viewpoints—two persistent failure modes observed in prior single-view texture generation approaches.


\subsection{User Study and Inference Efficiency}

We conduct a human evaluation study on 60 diverse test images to assess the quality of the generated results. Specifically, evaluators compare several methods across multiple dimensions, including geometry quality, texture quality, geometric rationality, texture rationality, image-to-3D consistency, and aesthetic quality. Here, geometry and texture quality measure the overall fidelity of the generated geometry and texture, such as structural completeness and richness of detail, whereas geometric rationality and texture rationality assess whether the reconstructed results remain plausible in regions not visible in the input image. Image-to-3D consistency reflects how well the generated textured mesh aligns with the input image, while aesthetic quality measures the overall visual appeal of both geometry and texture. As illustrated in Figure~\ref{fig:user_study_efficiency}, \methodName{} consistently receives higher ratings across all evaluated dimensions.

To evaluate the efficiency of our method, we compare inference time on a single GPU, with results reported in Figure~\ref{fig:user_study_efficiency}. As shown, our method requires substantially less inference time than existing DiT-based approaches. Notably, it is more efficient than the two-stage pipeline of TRELLIS-2.0, while also being faster than other single-stage methods.

%% file: sections/application.tex
\section{Applications}
\begin{figure}[htbp]
    \centering
    \setlength{\tabcolsep}{2pt}
    \renewcommand{\arraystretch}{0}
    \begin{tabular}{cc}
        \includegraphics[width=0.495\linewidth,height=0.22\textheight,keepaspectratio]{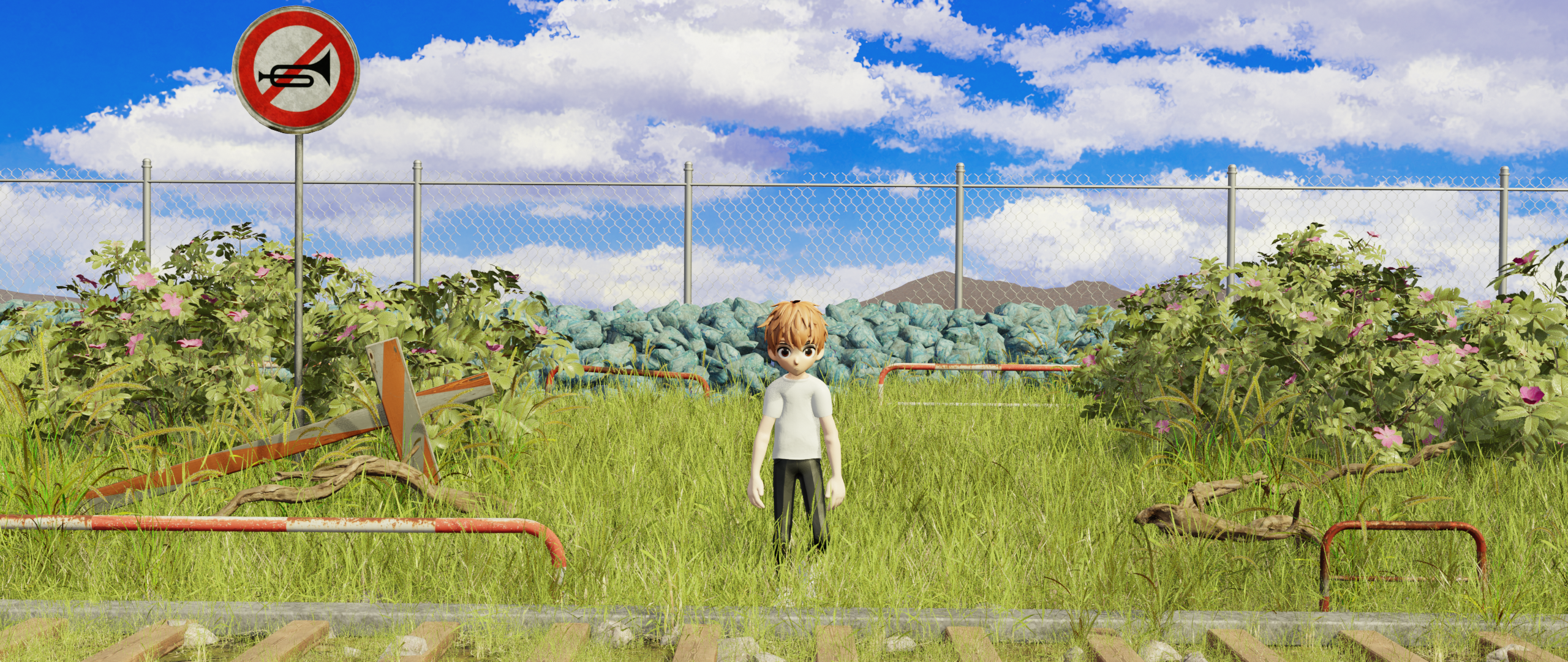} &
        \includegraphics[width=0.495\linewidth,height=0.22\textheight,keepaspectratio]{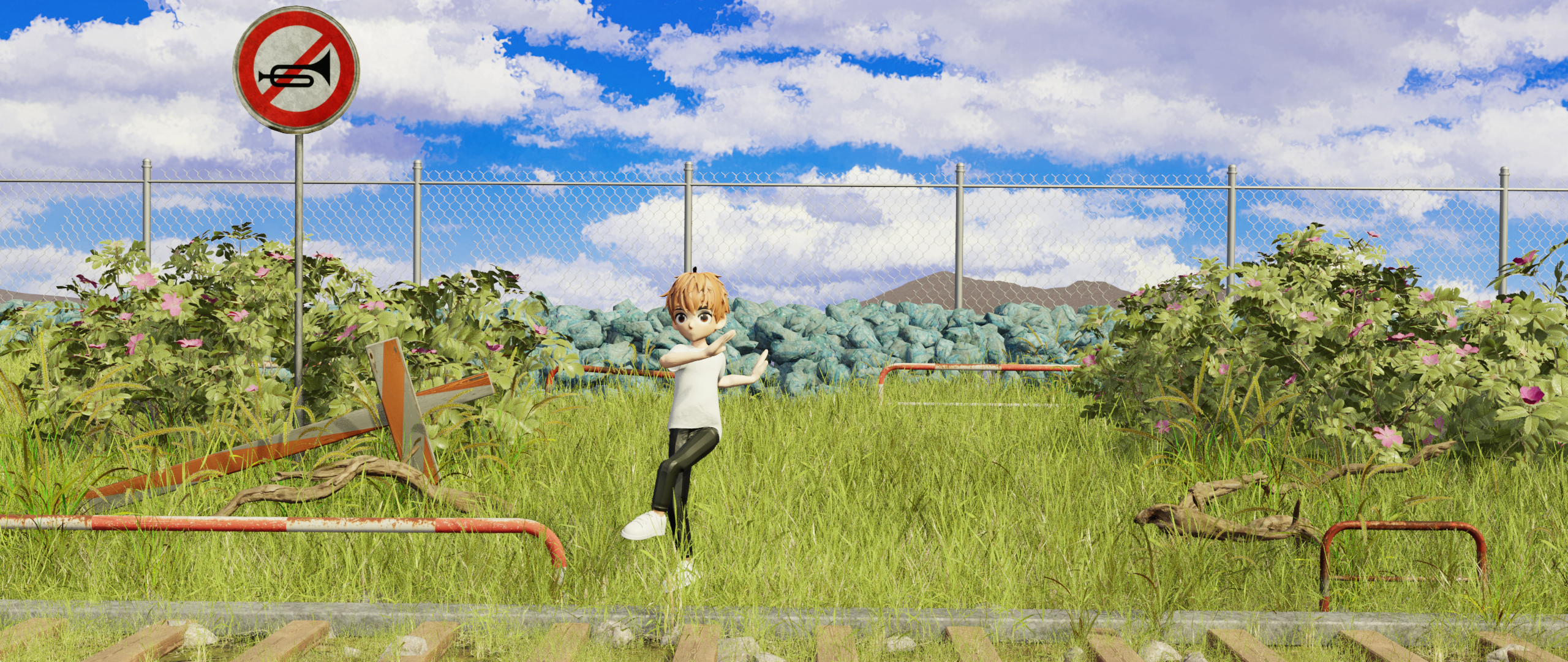} \\
    \end{tabular}
    \caption{\textbf{Animation-ready character generation results.} The rigged character exhibits natural deformation and consistent surface textures under different skeletal poses.}
    \label{fig:character_animation}
\end{figure}

Our framework’s core practical application lies in generating animation-ready 3D human characters from a single input reference image. As demonstrated in Figure~\ref{fig:character_animation}, given only one reference image, \methodName{} outputs a complete, fully textured character asset compatible with standard rigging and animation workflows out of the box. When driven by large-range articulated skeletal motions, the generated character maintains consistent realistic visual appearance, intact coherent geometry, and physically natural mesh deformations without obvious artifacts. These cases verify that our generated assets deliver high visual fidelity for static rendering while remaining stable and practical for downstream rigging, deformation, and motion editing tasks.

%% file: sections/limitation.tex
\section{Limitations}
Despite the strong performance and practical utility of \methodName{}, several limitations remain.

First, our training data still lacks sufficient diversity in character identities, clothing, accessories, poses, and styles, which may hinder generalization to rare or out-of-distribution cases. Scaling up the size and coverage of character data would likely further improve robustness and generalization.

Second, our geometry representation builds on SDFs, which inherently yield watertight meshes. While this benefits stability, rigging, and animation, it also complicates the modeling of non-watertight structures, open surfaces, and very thin details.

Third, the overall pipeline remains lengthy and depends on multiple models. Although this modular design improves stability and controllability, it inevitably increases system complexity. A promising direction is to move toward a more unified model, which would, however, likely demand larger-scale and higher-quality data.

Finally, the inference time is still noticeably longer than that of standard image generation models, limiting efficiency in practical deployment. Further optimization of both the pipeline and the model design~\cite{wang2026assetgen} is thus needed to reduce runtime.

Looking ahead, we believe it is promising to enrich the character data, explore more flexible geometric representations, streamline the multi-stage pipeline into a unified model, and improve runtime efficiency. We expect these directions to further enhance the quality, generalization, and practicality of our framework.

%% file: sections/conclusion.tex
\section{Conclusion}
In this work, we present \methodName{}, a unified image-conditioned 3D character generation framework optimized comprehensively for geometric fidelity, photorealistic texturing, native animation compatibility, and industrial inference efficiency. Built upon pre-trained 3D foundation models, our pipeline boosts character geometry via hierarchical coarse-to-fine structural refinement, aesthetic regularization, and multi-metric geometric preference reinforcement learning customized for human anatomical features and animatable properties. Such integrated capabilities cannot be attained by vanilla generic 3D foundation models or existing task-specific character generation paradigms. For texture modeling, our framework integrates multi-view texture back-projection, sparse-voxel 3D occlusion inpainting and semantic attribute decomposition to yield identity-consistent, view-stable character textures under complex lighting scenarios and severe self-occlusion. 

Apart from high-fidelity static 3D character reconstruction, \methodName{} natively supports full downstream animation pipelines, including video-guided facial expression transfer, skeletal motion retargeting, automatic mesh rigging, and weight-based skinning. This greatly enhances the practical applicability of generated assets for game design, virtual avatar creation, and interactive content production. Extensive quantitative benchmarks, qualitative visual comparisons, and human user studies validate that our framework surpasses existing state-of-the-art methods in both perceptual generation quality and production-oriented usability. We envision this work bridging the long-standing gap between academic 3D generative research and industrial asset production, facilitating the deployment of reliable, animation-friendly 3D character generation techniques into real-world creative workflows.

%% file: sections/appendix.tex
\section{Acknowledgments}

We thank Tingshuo Chen, Xiang Chen, Zhiguang Lu, Jiexi Wang, Zhongtian Yao, and Tongxin Zhang for their support in the development of the inference framework; Ziqian Chen, Kai Deng, Jing Huang, Lei Huang, Hao Li, Kuijiang Li, Guoqing Lian, Shuting Shen, Tong Wei, Zihang Xia, Chenrui Zhang, Longtao Zhang, Xinyi Zhang and Qinghua Zhou for their support with the data; and Li Wang, Xionghui Wang, Zihao Yu, Jianfeng Zhang, and Yifan Zhu for helpful discussions.